\documentclass[conference]{IEEEtran}
\usepackage{cite}
\IEEEoverridecommandlockouts
\usepackage{tikz}
\usetikzlibrary{arrows.meta,positioning,fit,backgrounds}
\usepackage{amsmath,amssymb,amsfonts}
\usepackage{algorithmic}
\usepackage{graphicx}
\usepackage{textcomp}
\usepackage{xcolor}
\usepackage[utf8]{inputenc} % allow utf-8 input
\usepackage[T1]{fontenc}    % use 8-bit T1 fonts
\usepackage{url}            % simple URL typesetting
\usepackage{booktabs}       % professional-quality tables
\usepackage{amsfonts}       % blackboard math symbols
\usepackage{amssymb}
\usepackage{amsthm}
\usepackage{algorithm}
\usepackage{nicefrac}       % compact symbols for 1/2, etc.
\usepackage{microtype}      % microtypography
\usepackage{xcolor}         % colors
\usepackage{subcaption}
\usepackage{mathtools}
\usepackage{siunitx}
\usepackage{svg}
\usepackage{enumitem}
\usepackage{adjustbox}
\usepackage{enumitem}
\usepackage{commath}
\usepackage{relsize}
\usepackage{tabularx}
\usepackage{cleveref}
\usepackage{bbm}
\usepackage{placeins}
\usepackage{multirow}
\usepackage[most]{tcolorbox}

\def\BibTeX{{\rm B\kern-.05em{\sc i\kern-.025em b}\kern-.08em
    T\kern-.1667em\lower.7ex\hbox{E}\kern-.125emX}}
\begin{document}

\title{Lifelong Representations: A Survey on Continual Self-Supervised Learning for Vision Models}

% \author{
% \IEEEauthorblockN{Anon}
% \and
% \IEEEauthorblockN{Anon}
% \and
% \IEEEauthorblockN{Anon}
% \and
% \IEEEauthorblockN{Anon}
% \and
% \IEEEauthorblockN{Anon}
% \and
% \IEEEauthorblockN{Anon}

\author{
    \IEEEauthorblockN{
        Sergi Masip\IEEEauthorrefmark{1}, 
        Alicja Dobrzeniecka\IEEEauthorrefmark{2}, 
        Jonathan Swinnen\IEEEauthorrefmark{1}, 
        Joachim Collin\IEEEauthorrefmark{1}, \\
        Bartłomiej Twardowski \IEEEauthorrefmark{3}, 
        Szymon Łukasik\IEEEauthorrefmark{2} and 
        Tinne Tuytelaars\IEEEauthorrefmark{1}
    }
    \vspace{0.15cm} % Optional: adds a tiny bit of breathing room
    
    \IEEEauthorblockA{\IEEEauthorrefmark{1}PSI, KU Leuven, Leuven, Belgium\\
    Email: \{sergi.masipcabeza, jonathan.swinnen, joachim.collin, tinne.tuytelaars\}@kuleuven.be}
    
    \IEEEauthorblockA{\IEEEauthorrefmark{2}NASK - National Research Institute, Warsaw, Poland \\
    Email: \{alicja.dobrzeniecka, szymon.lukasik\}@nask.pl}
    
    \IEEEauthorblockA{\IEEEauthorrefmark{3} Computer Vision Center, Universitat Autonoma de Barcelona, Barcelona, Spain \\
    IDEAS Research Institute, Warsaw, Poland \\
    Email: btwardowski@cvc.uab.cat}
}

% ONLY FOR PREPRINT
\newcommand{\mypostprintnotice}{%
  \begin{minipage}{\textwidth}
    \footnotesize
    \hrule\vspace{0.2cm}
    © 2026 IEEE. Personal use of this material is permitted. Permission from IEEE must be obtained for all other uses, in any current or future media, including reprinting/republishing this material for advertising or promotional purposes, creating new collective works, for resale or redistribution to servers or lists, or reuse of any copyrighted component of this work in other works. This work has been accepted for publication in IEEE EAIS 2026.
  \end{minipage}%
}
\makeatletter
\def\ps@IEEEtitlepagestyle{%
  \def\@oddfoot{\mypostprintnotice}%
  \def\@evenfoot{}%
}
\makeatother

\maketitle

\begin{abstract}
Traditionally, continual learning has assumed access to labeled data, yet many real-world applications---such as lifelong robotics---require models to adapt continuously from unlabeled streams. This has led to the development of continual self-supervised learning (CSSL), a rapidly growing area that lacks a dedicated, systematic review. In this work, we present a comprehensive survey of CSSL for vision, with connections to emerging vision-language settings. First, we analyze existing evaluation protocols and highlight inconsistencies that hinder fair comparison. We then examine why self-supervised objectives exhibit improved robustness to catastrophic forgetting, relating this to task-agnostic representations and smoother loss landscapes. Next, we organize existing methods into a unified taxonomy based on their forgetting-mitigation strategies, including distillation, replay, regularization, architectural approaches, model merging, and objective-level adaptation. Finally, we identify open challenges such as scalability and the need for fast adaptability.
We argue that advancing CSSL requires moving beyond small-scale benchmarks towards continual pre-training paradigms for large-scale systems.

\end{abstract}


\section{Introduction}\label{sec:intro}
% Sergi 2

% The emergence of large-scale foundation models has not resolved a fundamental problem in machine learning: how should a deployed model update itself as the world changes around it? 
% Imagine a delivery robot vision-language backbone trained on web-scale data from North America and Europe will encounter object appearances, road layouts, and lighting conditions largely absent from its training distribution when deployed in Sub-Saharan Africa~\cite{ramaswamy2023geode}. A medical imaging model trained on data from one hospital network faces a systematic distributional shift when extended to another. The industry standard is to periodically retrain or fine-tune models using updated data. However, both approaches have fundamental limitations at the scale of foundation models: retraining from scratch is computationally expensive, while fine-tuning does not provide a consistent way of accumulating knowledge over time, which often results in a deterioration of previously acquired capabilities. The alternative---continual adaptation---poses a challenge that has preoccupied the machine learning community for decades: how can new knowledge be incorporated without overwriting the old?
Most machine learning models are trained under a static paradigm, assuming the entire training dataset is available upfront. However, many real-world deployment scenarios challenge this assumption. Consider, for example, a robot navigation policy powered by a vision foundation model backbone that has been trained for logistics and delivery in North America and Europe. These environments are well represented in current datasets~\cite{ramaswamy2023geode}. However, when deployed in, for example, Sub-Saharan Africa, the model encounters new object appearances, road layouts and lighting conditions that were largely absent from its training data. The industry standard is to periodically retrain the model from scratch, which is wasteful. Yet the alternative--continually adapting the model--poses a fundamental challenge: how can new knowledge be incorporated without overwriting the old?

Continual Learning (CL) addresses this challenge by enabling models to learn from data as it arrives over time, while also reducing the risk of catastrophic forgetting.~\cite{MCCLOSKEY1989109}. However, most existing studies on CL focus on supervised settings, such as class-incremental learning~\cite{van2018three}, thereby sidestepping a key practical issue: in real-world deployments, labels are scarce, expensive, and often unavailable. A robot adapting to a new environment cannot pause to wait for human annotators. A foundation model ingesting a new data stream cannot assume curated, task-annotated inputs. Unlabeled visual data, by contrast, is generated at massive scales wherever these systems are deployed. Self-supervised learning, which extracts supervisory signals from the structure of the data itself~\cite{jing2021selfsupervisedlearning}, is therefore a natural alternative for continual adaptation of models in the wild.

\begin{figure}[t]
    \centering
    \includegraphics[width=0.4\textwidth]{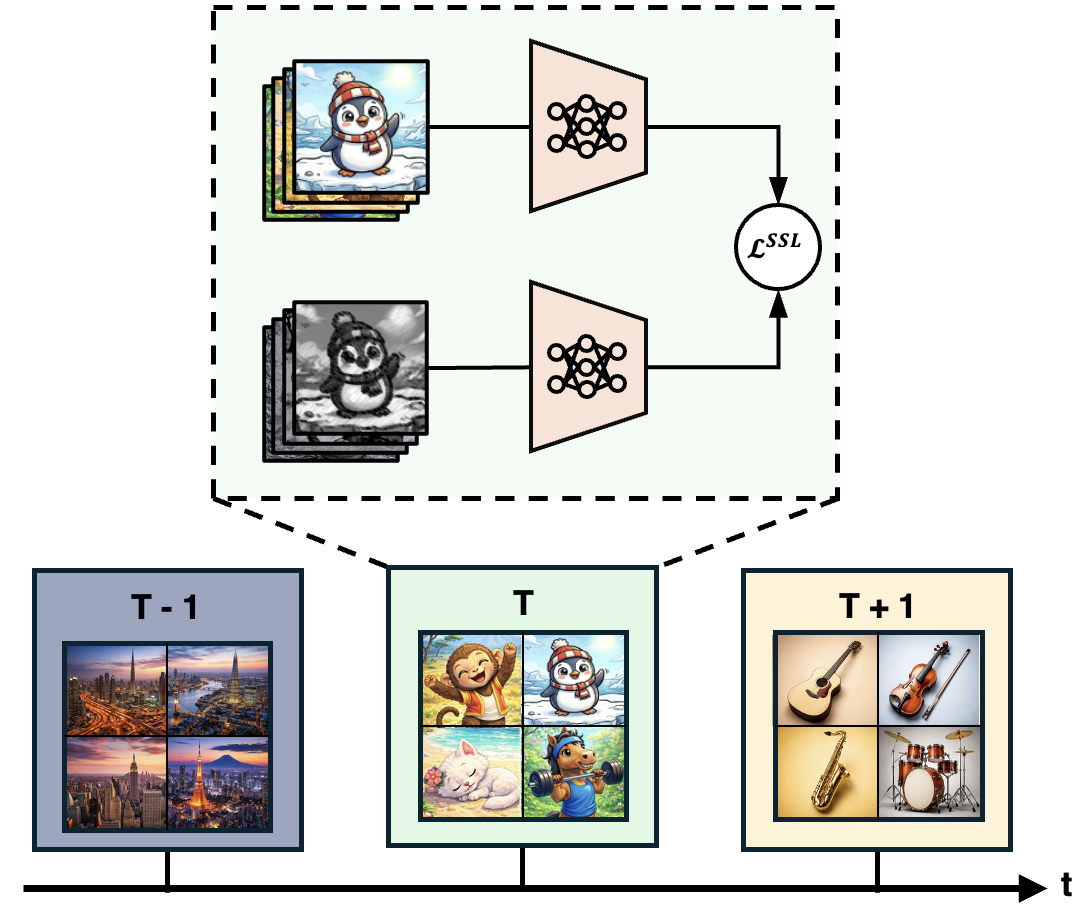} 
    \caption{A conceptual visualization of CSSL. The bottom part represents tasks $T$ arriving in sequence, and the top part shows an illustration of an SSL objective (in this example, non-contrastive).}
    \label{fig:cssl_example}
\end{figure}

This has driven growing interest in continual self-supervised learning (CSSL), in which a model is continually pre-trained as new streams of data arrive, without explicit labels (\Cref{fig:cssl_example}). This setting remains underexplored relative to its supervised counterpart and is almost entirely unexplored at foundation-model scale~\cite{bell2025futurecontinuallearningera}. % While self-supervised and continual learning are well-surveyed independently~\cite{lange2022cl_survey}, their intersection lacks a dedicated review.  
While continual learning has been surveyed for large language models~\cite{shi2025cl_and_llms} and for the finetuning of vision-language models (VLMs)~\cite{liu2025vlm_cl_survey}, CSSL in vision and VLMs lacks a dedicated review. The closest works~\cite{bagus2022supervisedcontinuallearningreview, kilickaya2023labelefficientincrementallearningsurvey} treat CSSL only as a secondary topic and do not discuss concrete protocols, metrics, benchmarks, theory, or recent advances. Because modern applications increasingly rely on multimodal streams rather than isolated unimodal encoders, restricting our analysis to pure vision would ignore the reality of deployed systems. To fill this gap, we present the first systematic review of CSSL for vision models, while naturally extending our analysis to vision-language models (VLMs) where multimodal streams intersect with visual representations. Our contributions are as follows:

%\begin{itemize}
%    \item \textbf{Evaluation analysis.} We gather and review typical evaluation protocols, benchmarks, and metrics. %, highlighting inconsistencies and gaps that hinder fair comparison across methods.
%    \item \textbf{Insight into forgetting dynamics in self-supervision.} We examine why self-supervised objectives appear more robust to catastrophic forgetting than supervised ones. % , and discuss the implications for scalable, lifelong adaptation of foundation models.
%    \item \textbf{Taxonomy of methods and open challenges.} We categorize existing approaches by their forgetting-mitigation strategies, identify the key challenges that must be addressed, and discuss how these approaches can be scaled to larger foundation model regimes.
%\end{itemize}

\begin{itemize}
    \item \textbf{Evaluation analysis.} We gather and review typical evaluation protocols, benchmarks, and metrics (\Cref{sec:setup}). %, highlighting inconsistencies and gaps that hinder fair comparison across methods.

    \item \textbf{Insight into forgetting in SSL.} We examine why self-supervised objectives appear more robust to catastrophic forgetting than supervised ones (\Cref{sec:properties}). % , and discuss the implications for scalable, lifelong adaptation of foundation models.

    \item \textbf{Taxonomy of methods.} We categorize CSSL approaches by their forgetting-mitigation strategies (\Cref{sec:cssl}). Additionally, we examine how SSL can be a useful component within hybrid, supervised CL pipelines (\Cref{sec:ssl_component}).

    \item \textbf{Identifying open questions.} We finally identify the key challenges that must still be addressed, and discuss how these approaches can be scaled to larger foundation model regimes (\Cref{sec:discussion}).

\end{itemize}

With this work, our aim is not only to map the current landscape, but also to position CSSL as a promising foundation for lifelong, large-scale, deployed systems.

\section{Objectives, Protocols and Evaluation}\label{sec:setup}
This section summarizes the main families of self-supervised objectives, training protocols, and evaluation metrics.

% -------------------------------------------------------

\subsection{Self-supervised learning objectives}
\label{subsec:ssl_objectives}

Self-supervised learning methods train an encoder to produce useful 
representations without requiring manual labels by solving \emph{pretext tasks} 
whose supervisory signal is derived directly from the data~\cite{jing2021selfsupervisedlearning}. Modern SSL approaches are commonly categorized into
contrastive, non-contrastive, and masked modeling methods,
which correspond broadly to contrastive, predictive,
and generative paradigms in prior surveys \cite{khan2025surveyself}. In our survey, we focus on continual learning with objectives within these families.

\paragraph{Contrastive learning}
Contrastive methods learn representations by aligning augmented views of the same image (positive pairs) while distinguishing them from different samples (negative pairs). These approaches were popularized by SimCLR~\cite{chen2020sim_clr} and MoCo~\cite{He2019MomentumCF,chen2020mocov2} with the InfoNCE objective~\cite{oord2018info_nce}. %, these approaches encode two augmented views and optimize their representations to be similar while distinguishing them from negative samples.

% Contrastive methods align augmented views of the same image while repelling different images (negative samples). 

% Popularized by SimCLR~\cite{chen2020sim_clr} and MoCo~\cite{He2019MomentumCF} using the InfoNCE objective~\cite{oord2018info_nce}, these frameworks map two views $\mathbf{x}_1^+$ and $\mathbf{x}_2^+$ via an encoder $f_\theta$ and projection head $g_\phi$ to embeddings $\mathbf{z}_1$ and $\mathbf{z}_2$, optimizing them for similarity against negatives.

\paragraph{Non-contrastive learning}
Explicit negatives for alignment are not strictly necessary to learn useful representations. Non-contrastive approaches avoid collapse via alternative mechanisms: self-distillation methods (BYOL~\cite{grill2020bootstrap_latent}, SimSiam~\cite{chen2022siamese_repr}) match views using asymmetric predictors and stop-gradients; redundancy-reduction techniques impose statistical constraints, such as minimizing cross-correlation off-diagonals (Barlow Twins~\cite{zbontar2021barlow_twins}) or decorrelating features (VICReg~\cite{bardes2022vic_reg}); and clustering methods (SwAV~\cite{caron2020cluster_learning}) map representations to learnable prototypes to predict one view's cluster assignment from another.

% Several subsequent works have shown that explicit negative samples are not strictly necessary for learning useful representations. Non-contrastive approaches avoid representation collapse through alternative mechanisms such as architectural asymmetry, statistical constraints, or clustering. Distillation-based architectures like BYOL~\cite{grill2020bootstrap_latent} and SimSiam~\cite{chen2022siamese_repr} rely on asymmetric predictor networks and stop-gradient operations to match representations across views. % For example, the objective often minimizes the negative cosine similarity between a prediction $\mathbf{p}_1 = h_\psi(\mathbf{z}_1)$ and the stop-gradient target $\mathbf{z}_2$:
% % \begin{equation}
% %     \mathcal{L}_{\mathrm{Distillation}} = -\frac{\mathbf{p}_1^\top \mathrm{sg}(\mathbf{z}_2)}{\|\mathbf{p}_1\|_2 \|\mathrm{sg}(\mathbf{z}_2)\|_2}
% % \end{equation}
% Redundancy-reduction methods instead impose statistical constraints: Barlow Twins~\cite{zbontar2021barlow_twins} minimizes the off-diagonal entries of the cross-correlation matrix between representations, while VICReg~\cite{bardes2022vic_reg} explicitly maintains variance and decorrelates feature dimensions. Finally, cluster-assignment methods like SwAV~\cite{caron2020cluster_learning} avoid explicit pairwise comparisons by mapping representations to soft cluster assignments via learnable prototypes, training the model to predict the cluster assignment of one view from the representation of another.

\paragraph{Masked modeling}
Inspired by NLP, masked image modeling (MIM) captures global structure by reconstructing missing parts of the input. Masked Autoencoders (MAE)~\cite{he2022masked_autoencoders} process visible patches via a ViT encoder, training a lightweight decoder to reconstruct masked regions. % Unlike contrastive methods, MIM representations typically excel under fine-tuning rather than linear probing~\cite{he2022masked_autoencoders}.

\subsection{Training protocols}
\label{subsec:protocols}

We define the training protocols based on whether task boundaries are defined and how.

\paragraph{Offline CSSL}
The standard evaluation protocol in continual self-supervised learning. Data arrive as a sequence of discrete experiences, each presented in full before the next begins. Task boundaries are typically explicit, indicating when a new experience starts, and models are allowed to train on this experience for either a set amount of epochs or until convergence. Most methods in the literature adopt this setting and evaluate representation quality after each experience. % using metrics such as linear-probe or $k$-NN accuracy.

\paragraph{Online / task-free CSSL}
In online CSSL, data arrive as a single stream of mini-batches, typically processed in a one-pass fashion without revisiting samples. Task identities are not provided, and boundaries may be absent or blurred. This setting introduces additional computational constraints and is therefore more challenging. The computational budget can be measured with, for example, the Cumulative Backward Passes (CBP)~\cite{cignoni2025clalatentalignmentonline} metric, a compute-normalized metric for comparing methods under strict online training budgets.

% ------------------------------------------------------------
\subsection{Evaluation metrics}

% The evaluation of a self-supervised model's representations during continual learning typically involves either probing them or directly comparing them across tasks through CKA~\cite{kornblith2019similarity}. The \emph{probe type} matters: linear probing measures linear separability of the representation, while $k$-NN accuracy evaluates the extent to which samples from the same class form coherent local neighborhoods in the embedding space. Let $a_{t,k}$ be the accuracy of the model obtained through probing after training on task $t$, evaluated on task $k$.

Evaluating CSSL typically involves either probing representations or analyzing their structural stability. The most common probe types are linear probing, which measures linear separability, and $k$-NN accuracy, which evaluates whether samples from the same class form coherent local neighborhoods. %Let $a_{t,k}$ be the probe accuracy on task $k$ after training on task $t$.

\paragraph{Standard CL metrics}
\textbf{Average Accuracy (AA)} %($\frac{1}{T} \sum_k a_{T,k}$) 
measures the final mean performance across all $T$ tasks. \textbf{Forgetting ($\mathbf{\mathcal{F}}$)} and \textbf{Backward Transfer (BWT)} evaluate the impact of learning new tasks on past tasks; $\mathcal{F}$ captures strict performance degradation, while signed BWT captures both interference and beneficial transfer. Conversely, \textbf{Forward Transfer (FWT)} measures how prior knowledge facilitates learning new tasks.

\paragraph{Multimodal metrics} In large-scale vision-language settings, metrics like \textbf{Average Knowledge Accumulation (AKA)} and \textbf{Average Zero-Shot Retention (AZS)} jointly capture the plasticity--stability trade-off across held-in and held-out task sets~\cite{udandarao2024fomo_flux, garg2024ticclip }. Both metrics are typically derived from accuracy and retrieval performance.

\paragraph{Representation stability (CKA)} To diagnose \emph{where} in the network forgetting occurs~\cite{Liu2025Branch}, several works analyze the learned representation space using \textbf{Centered Kernel Alignment (CKA)}~\cite{kornblith2019similarity}. CKA measures the similarity of activation Gram matrices between models or checkpoints; values near $1$ indicate nearly identical representations up to orthogonal transformation, while values near $0$ indicate substantial divergence.

% \paragraph{Multimodal metrics}
% In multimodal and large-model settings, \emph{Average Knowledge Accumulation} (AKA) and \emph{Average Zero-Shot Retention} (AZS) jointly capture the plasticity--stability trade-off across held-in and held-out task sets~\cite{udandarao2024fomo_flux, garg2024ticclip}.

% ------------------------------------------------------------
\begin{table}[t]
\centering
\caption{Common benchmarks used in continual self-supervised learning, grouped by evaluation setting.}
\label{tab:benchmarks}
\setlength{\tabcolsep}{3pt}
\begin{tabular}{p{3.6cm} p{1.6cm} p{2cm} p{0.8cm}}
\toprule
\textbf{Benchmark} & \textbf{Protocol} & \textbf{Modality} & \textbf{Scale} \\
\midrule

\multicolumn{4}{l}{\textbf{Vision Benchmarks}} \\
\midrule
CIFAR-10/100~\cite{krizhevsky2009learning} & Offline/Online & Vision & Small \\
ImageNet-100~\cite{deng2009imagenet} & Offline/Online & Vision & Medium \\
TinyImageNet~\cite{deng2009imagenet} & Offline/Online & Vision & Medium \\
% SCALE & Online & Vision & Small \\ % Medium \\

\midrule
\multicolumn{4}{l}{\textbf{Multimodal Benchmarks}} \\
\midrule
% P9D\cite{zhu2023ctp} & Offline & Vision-Language & Large \\
TiC-(DataComp, YFCC, \newline RedCaps)~\cite{garg2024ticclip} & Offline & Vision-Language & Large \\
FoMo-in-Flux~\cite{udandarao2024fomo_flux} & Offline & Vision-Language & Large \\

\bottomrule
\end{tabular}
\end{table}

\subsection{Benchmarks}
\label{subsec:benchmarks}

The most widely used benchmarks split standard classification datasets into a sequence of tasks by class. In Table \ref{tab:benchmarks}, we have classified the most common benchmarks based on the modality they evaluate. % Say that some works use domain-specific but we omit them.

\section{Properties of SSL in continual learning setups}\label{sec:properties}
% first draft below vv, feel free to change stuff if you find inaccuracies or if there is something you don't like :)
A recurrent observation when running self-supervised methods in a continual learning setup is that the representations seem to be more robust to forgetting than when using supervised losses~\cite{madaanrepresentational, ni2021selfsupervisedclassincrementallearning, fini2022cassle}. Studies have empirically shown, using CKA, that the representations remain remarkably stable throughout training. Before examining the continual learning approaches applied to CSSL, we review why self-supervised learning is less susceptible to catastrophic forgetting per se.

\subsection{Explaining the robustness of SSL representations}

% Empirical and analytical studies have attributed this robustness to a series of factors: the properties of SSL representations and the geometry of the loss landscape. Finally, we identify the main challenges in continual self-supervised learning.

Empirical and analytical studies suggest that the robustness of SSL models to forgetting stems from two core properties: their representations are task-agnostic, and their loss landscapes are smoother and flatter. This perspective clarifies why forgetting is less problematic in CSSL and highlights the key challenges that remain.

\paragraph{Supervision, representation collapse, and tunnel effect}
During supervised classification training, representations often suffer from a loss of informational richness, manifesting in two related phenomena. First, representations often exhibit neural collapse, where samples of the same class converge to a single point in representation space, and class means form a highly symmetric structure \cite{papyan2020prevalence_collapse,tirer2023perturbation_collapse}. While this geometry improves linear separability for the training task, it suppresses within-class variability and discards features that may be useful for other tasks \cite{kornblith2021transfer}. Second, a related phenomenon, known as the tunnel effect~\cite{masarczyk2023tunnel_effect}, has also been observed in supervised models, where deeper representations collapse to a progressively lower-rank subspace. This effect, which worsens when training with few classes (as is common in traditional class-incremental learning), reduces out-of-distribution accuracy and causes significant forgetting in continual learning~\cite{harun2024variables_generalization}.

\paragraph{Self-supervision learns task-agnostic features}
On the other hand, SSL objectives encourage the network to learn task-agnostic, descriptive features about the data distribution itself~\cite{ozbulakknow}. Self-supervised representation geometry seems to have higher effective dimensionality and less class‑wise collapse than supervised models~\cite{ben2023reverse, song2025discriminability}. In fact, many SSL methods are explicitly designed to prevent representational collapse during training~\cite{zbontar2021barlow_twins, bardes2022vic_reg, caron2021dino, chen2022siamese_repr, assran2023jepa, grill2020bootstrap_latent}, and a higher rank tends to correlate with better downstream performance~\cite{garrido2023rankme}.

% However, a recent study~\cite{chung2026global} argues that avoiding collapse might not be a guarantee of representational quality; instead, they propose a new metric called the Jacobian Effective Rank, which measures how many independent input directions the model is sensitive to.
% TODO: Revise paragraph above. Some papers indicate nuances and weaknesses of SSL worth discussing.

% Maybe mention paper on sequential training on imagenet (HOW WELL DOES SELF-SUPERVISED PRE-TRAINING PERFORM WITH STREAMING DATA?) and CLA where they show that SSL can be competitive to joint pre-training. 

\paragraph{Flatter minima in the loss landscape}
The robustness to forgetting can also be explained by looking at the loss landscape. Several works~\cite{huwell, madaanrepresentational, bhat2022task} have observed that self-supervised objectives sculpt a smoother loss landscape compared to supervised objectives. On top of this, the solutions reached by these objectives are generally flatter, which has benefits against forgetting, as it increases the tolerance to parameter shifts.

% Maybe discuss this paper: Must Unsupervised Continual Learning Relies on Previous Information? (CVPRW 2024) 37

\subsection{The challenges of continual self-supervised learning}\label{subsec:challenges}

\paragraph{Data and compute hunger}
A first limitation shared with SSL in general is that these methods are data-hungry, requiring more iterations and data to converge than their supervised counterparts~\cite{addepalli2022towards, koccyiugit2023accelerating, imai2024faster}. This is further aggravated in the continual setting, where small per-task datasets have been shown to cause significant instability~\cite{fini2022cassle}. Compounding this, SSL carries a substantial computational overhead: long training schedules, large batch sizes, and auxiliary components such as momentum encoders. % or projection heads.

% \paragraph{Sensitivity to hyperparameters}
% A related practical concern is the sensitivity of SSL methods to hyperparameters and augmentation strategies. Contrastive and self-distillation methods are known to be sensitive to design choices such as crop ratios and color jitter strength, and these choices may not remain optimal as the data distribution shifts across tasks, adding a non-trivial tuning burden. % TODO: references

\paragraph{Stability vs. plasticity dilemma}
Regarding continual learning, catastrophic forgetting remains an open problem. The performance gap with respect to joint training---commonly used as an upper bound---is still present and has been observed to widen as the number of tasks grows~\cite{ni2021selfsupervisedclassincrementallearning, fini2022cassle, gomez2022cl_pfr, gomez2024plasticity}. At the same time, several works argue that plasticity deserves at least equal attention as a research focus~\cite{gomez2024plasticity, cossu2024continual}, as the two objectives are inherently in tension and the absence of a supervised signal makes balancing the stability-plasticity trade-off particularly challenging in CSSL.

\paragraph{Risk of overfitting}
Rehearsal-based methods (see \Cref{sec:replay_methods}) face an additional challenge: because SSL requires far more training epochs than supervised learning, memory samples are revisited many more times, increasing the risk of overfitting to the replay buffer and undermining its purpose~\cite{wu2023adaptive, fini2022cassle}.

\paragraph{Cross-modal drift in multimodal settings}
In multimodal continual learning, particularly for vision–language models, an additional challenge arises from maintaining the alignment of modalities over time. As models are updated with new data, the shared embedding space between modalities can drift. This leads to a decline in cross-modal retrieval performance and zero-shot generalization \cite{udandarao2024fomo_flux, garg2024ticclip}. Unlike in unimodal settings, where forgetting primarily affects feature representations, multimodal models must preserve both intra-modal structure and inter-modal consistency.

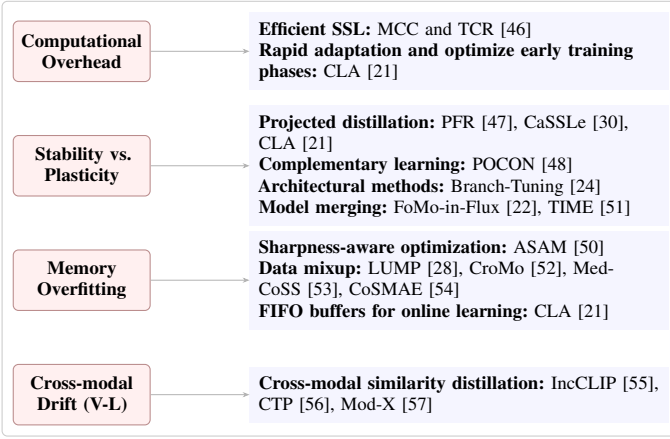
\begin{figure}
\centering
\begin{tikzpicture}[
  font=\scriptsize,
  node distance=0.7cm and 1.3cm,
  cbox/.style={
    rectangle, rounded corners=2pt,
    draw=red!40!black!45, fill=red!6!white,
    text width=1.6cm, align=center,
    minimum height=0.8cm,
    inner sep=3pt,
    font=\scriptsize\bfseries
  },
  sbox/.style={
    rectangle,
    draw=none, fill=blue!4!white,
    text width=5.3cm, align=left,
    inner sep=3.5pt,
    font=\scriptsize
  },
  sep/.style={draw=black!15, thin},
  arr/.style={-{Stealth[scale=0.55]}, very thin, draw=black!30},
]

% Row 1
\node[cbox] (c1) {Computational\\Overhead};
\node[sbox, right=of c1] (s1) {
\textbf{Efficient SSL:} MCC and TCR~\cite{imai2024faster}\\
\textbf{Rapid adaptation and optimize early training phases:} CLA~\cite{cignoni2025clalatentalignmentonline}
};
\draw[arr] (c1.east) -- (s1.west);

% Row 2
\node[cbox, below=of c1] (c2) {Stability vs.\\Plasticity};
\node[sbox, right=of c2] (s2) {
\textbf{Projected distillation:} PFR~\cite{gomez2022cl_pfr}, CaSSLe~\cite{fini2022cassle}, CLA~\cite{cignoni2025clalatentalignmentonline}\\
\textbf{Complementary learning:} POCON~\cite{gomez2024plasticity}\\
\textbf{Architectural methods:} Branch-Tuning~\cite{Liu2025Branch}\\
\textbf{Model merging:} FoMo-in-Flux~\cite{udandarao2024fomo_flux}, TIME~\cite{dziadzio2025merge}
};
\draw[arr] (c2.east) -- (s2.west);

% Row 3
\node[cbox, below=of c2] (c3) {Memory\\Overfitting};
\node[sbox, right=of c3] (s3) {
\textbf{Sharpness-aware optimization:} ASAM~\cite{wu2023adaptive}\\
\textbf{Data mixup:} LUMP~\cite{madaanrepresentational}, CroMo~\cite{mushtaq2024cromo}, MedCoSS~\cite{ye2024medcoss}, CoSMAE~\cite{mollenbrok2025cosmae}\\
\textbf{FIFO buffers for online learning:} CLA~\cite{cignoni2025clalatentalignmentonline}
};
\draw[arr] (c3.east) -- (s3.west);

% Row 4
\node[cbox, below=of c3] (c4) {Cross-modal\\Drift (V-L)};
\node[sbox, right=of c4] (s4) {
\textbf{Cross-modal similarity distillation:} IncCLIP~\cite{yan2022gntr_vlp}, CTP~\cite{zhu2023ctp}, Mod-X~\cite{ni2023modx}
};
\draw[arr] (c4.east) -- (s4.west);

% Background + separators
\begin{scope}[on background layer]
  \draw[sep] ([yshift=-0.2cm]c1.south -| s1.west) -- ([yshift=-0.2cm]c1.south -| s1.east);
  \draw[sep] ([yshift=-0.2cm]c2.south -| s1.west) -- ([yshift=-0.2cm]c2.south -| s1.east);
  \draw[sep] ([yshift=-0.2cm]c3.south -| s1.west) -- ([yshift=-0.2cm]c3.south -| s1.east);

  \node[draw=black!20, rounded corners=2pt, fill=white,
        fit=(c1)(c4)(s1)(s4), inner sep=4pt] {};
\end{scope}

\end{tikzpicture}
\caption{Challenges in CSSL and representative solutions.}
\label{fig:cssl_challenges}
\end{figure}

Figure \ref{fig:cssl_challenges} summarises these key challenges and highlights the methods proposed to address each of them.

\section{Continual Self-Supervised Learning}\label{sec:cssl}
\begin{figure}[t]
    \centering
    \includegraphics[width=0.45\textwidth]{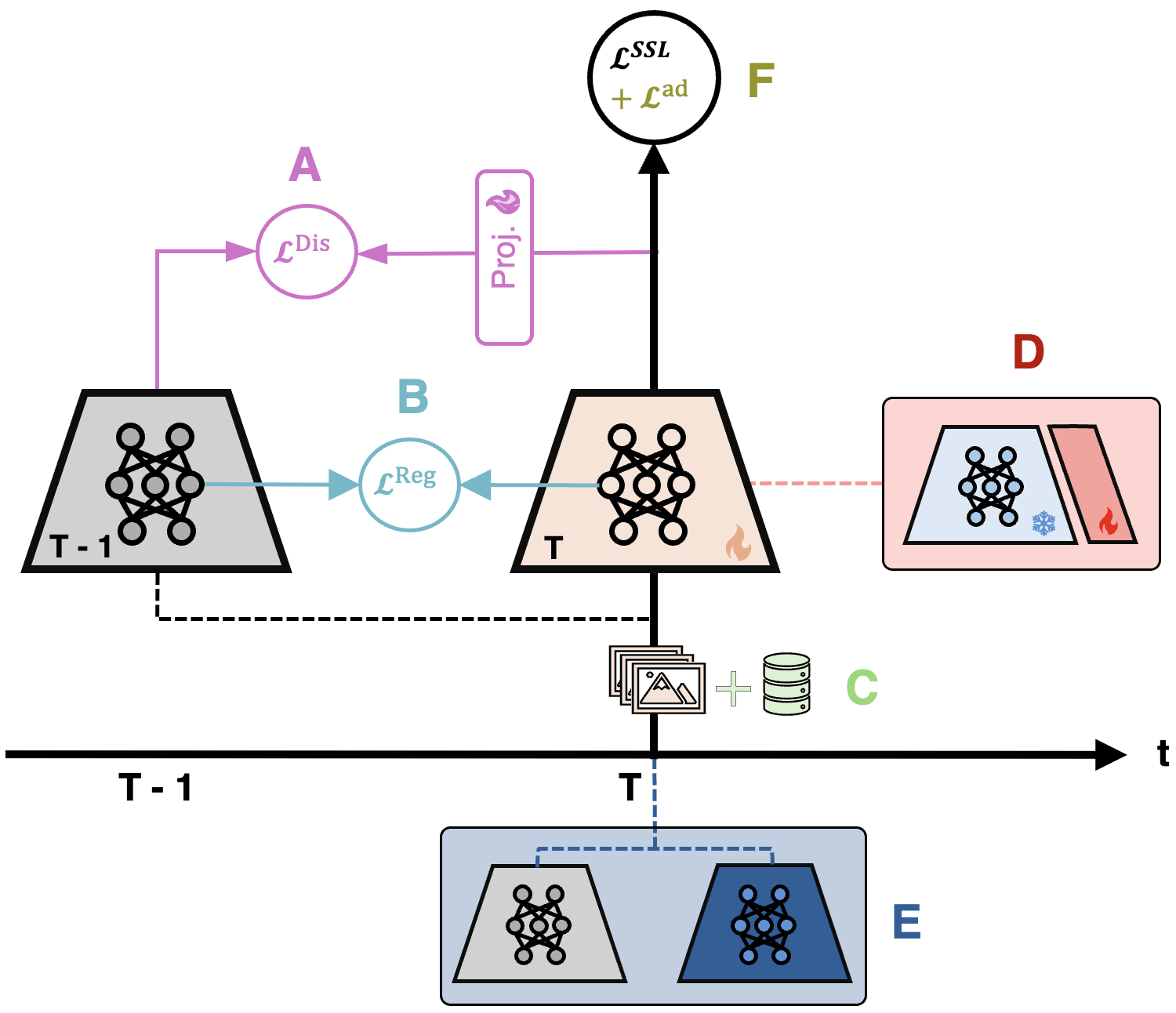} 
    \caption{The diagram illustrates the different approaches to CSSL: (A) A distillation loss ($\mathcal{L^{\text{Dis}}}$) is used to align representations of the current model with those of the previous one ($T-1$) with an optional projector; (B) a weight regularization loss ($\mathcal{L^{\text{Reg}}}$) penalizes changes to parameters that were important for past tasks, computed by comparing the current model ($T$) against the frozen previous model ($T-1$); (C) incoming data at step $T$ consists of new images combined with a replay memory buffer of past samples; (D) a parameter-isolated architecture separates frozen and trainable components, enabling task-specific plasticity while preserving prior knowledge; (E) a model merging module combines the previous and current model checkpoints; (F) the overall training objective optionally adapts ($\mathcal{L^{\text{ad}}}$) the self-supervised loss ($\mathcal{L^{\text{SSL}}}$) to embed continual learning constraints directly into the SSL objective.}
    \label{fig:taxonomy}
\end{figure}

CSSL methods aim to maintain stable and transferable representations while expanding the model's knowledge. As discussed in \Cref{sec:properties}, despite increased representational stability, models still suffer from gradual forgetting and loss of plasticity when trained through CSSL. In this section, we gather and discuss papers that are strictly self-supervised, i.e., no labels were used during training. We identify six main families of approaches: distillation, weight regularization, replay, architectural methods, model merging, and objective-level adaptation. We illustrate the different approaches in \Cref{fig:taxonomy}.

\subsection{Distillation}

Knowledge distillation has emerged as a promising strategy for mitigating catastrophic forgetting in CSSL. Traditionally, distillation aimed to maintain stability by preserving the feature structure learned in previous tasks. 

MedCoSS \cite{ye2024medcoss} explored training a shared MAE encoder sequentially on different medical data modalities, using a distillation loss on representations of samples from past tasks.
Contrastive Continual Learning (CCL) \cite{lin2022continual}, on the other hand, distills similarity matrices of sample representations directly from a momentum teacher network. They found that this improves performance, although the gains are modest in general. In vision-language domains, distillation is commonly used to preserve cross-modal alignment over time. Most approaches adopt \textit{cross-modal similarity distillation}, matching the softmax-normalized similarity distributions of a previous model using KL divergence on both image-to-text and text-to-image pairs~\cite{zhu2023ctp, yan2022gntr_vlp, garg2024ticclip}, thereby preserving the shared embedding space structure and mitigating cross-modal drift. Mod-X~\cite{ni2023modx} operates directly on non-normalized similarity matrices to enforce strong geometric consistency. CTP~\cite{zhu2023ctp} extends similarity distillation to intra-modal pairs to preserve modal structure, and adds a contrastive loss between the current model and an EMA teacher to stabilize representation drift.
% 

% distillation aims to preserve both the structure and the cross-modal alignment of the shared embedding space across time. A common method is to naïvely match the softmax-normalized similarity distributions of the previous model using KL divergence on both image-to-text and text-to-image pairs~\cite{zhu2023ctp, yan2022gntr_vlp, garg2024ticclip}, sometimes extended to intra-modal pairs~\cite{zhu2023ctp}. Alternatively, distillation can be integrated directly into contrastive learning; for example, CTP~\cite{zhu2023ctp} introduces a compatible momentum teacher to guide unimodal contrastive encoder updates, alongside the distillation of masked language modeling outputs and topology preservation.

% In multimodal settings, distillation is commonly used to preserve cross-modal alignment over time. Most approaches match the softmax-normalized similarity distributions of the previous model using KL divergence on both image-to-text and text-to-image pairs, thereby maintaining the joint embedding structure. Mod-X operates on non-normalized similarity matrices to enforce stronger geometric consistency. CTP extends similarity distillation to intra-modal pairs in order to preserve modal structure, and performs contrastive learning between the current model and an EMA teacher to encourage slow unimodal drift and stabilize representation updates.

In unimodal vision settings, it has been observed that naive distillation, where features are matched directly, severely restricts model plasticity by anchoring representations too rigidly to the past, stripping the model's ability to adapt to new data distributions~\cite{gomez2022cl_pfr, fini2022cassle, cignoni2025clalatentalignmentonline}. To overcome this limitation, recent methodologies have adopted \textit{projected distillation}, incorporating an auxiliary projector or predictor network into the distillation process, which acts as a buffer against distribution shifts~\cite{ni2021selfsupervisedclassincrementallearning}. This preserves the encoder's plasticity as long as new representations can be mapped back to older ones~\cite{gomez2022cl_pfr}, a representational benefit significant enough even to improve standard supervised baselines~\cite{marczak2024revisiting}.

Pioneering frameworks such as CaSSLe~\cite{fini2022cassle} and Projected Functional Regularization (PFR)~\cite{gomez2022cl_pfr} map current representations back to the previous task model via a projector, thereby explicitly freeing the encoder to learn new features. Building on CaSSLe, Pseudo-Negative Regularization (PNR)~\cite{cha2024regularizing} adapts projected distillation by incorporating the current model's outputs as pseudo-negatives, explicitly preventing new representations from overlapping with previously learned ones. Extensions like Plasticity-Optimized Complementary Networks~\cite{gomez2024plasticity} apply this to transfer knowledge from a task-specific feature extractor to a cumulative knowledge extractor in a complementary learning system. One limitation is that these methods depend on task boundaries. This motivates Continuous Latent Alignment (CLA)~\cite{cignoni2025clalatentalignmentonline} to adapt the CaSSLe framework to the online setting by replacing the previous task model with an EMA of the model. However, CLA scales poorly with increased compute and primarily benefits early pretraining, acting more as a representational regularizer than a complete continual learning solution.

Projected distillation has also been adapted across different modalities. For example, in remote sensing, CoSMAE~\cite{mollenbrok2025cosmae} continually pretrains an MAE while applying a model mixup strategy, interpolating between the current and previous models, and aligning the current model with this mixed teacher via a projector. In vision-language domains, C-CLIP~\cite{liu2025cclip} performs contrastive distillation in a projected joint embedding space to avoid over-constraining the updated features.
%In mask modeling approaches, CoSMAE~\cite{mollenbrok2025cosmae} applies a model mixup strategy, aligning the outputs of the current model with this mixed target via a projector. In remote sensing, recent work~\cite{mollenbrok2025plasticity} enhances projected distillation by decoupling the projected features into task-common and task-specific components. To balance memory stability and learning plasticity, the task-common features are correlated, whereas the task-specific features are decorrelated. 

\textbf{Takeaway:}
These works demonstrate that augmenting distillation with a projector is a highly effective approach for balancing stability and plasticity in CSSL. However, practitioners must note that this family of approaches increases the computational overhead by adding a teacher network and a projector, requiring at a minimum one additional forward pass. Improving the efficiency of these methods is an open question.

\subsection{Weight regularization}
Weight regularization techniques, such as EWC~\cite{kirkpatrick2017overcoming_forgetting}, SI~\cite{zenke2017synaptic}, and MAS~\cite{aljundi2018mas}, mitigate catastrophic forgetting by penalizing changes to parameters deemed critical for previous tasks. %In CSSL, these methods are directly integrated into self-supervised objectives to stabilize models.

EWC is frequently utilized to consolidate weights during unsupervised incremental phases. For instance, VINIL~\cite{kilickaya2023labels} combines EWC with memory replay to manage long-horizon visual instance learning without manual labels. In remote sensing, Continual Barlow Twins~\cite{marsocci2023cbt} trains a Barlow Twins backbone \cite{zbontar2021barlow_twins} and employs EWC to preserve representations during incremental pretraining on heterogeneous earth observation datasets. %, cutting computational time by nearly 50\% compared to joint training with negligible performance drops. 
%EWC is also used as a supplementary regularizer alongside projected feature distillation in methods like PFR~\cite{gomez2022cl_pfr, fini2022cassle}. 
SI performs particularly well when integrated with unsupervised visual representation methods like SimSiam and Barlow Twins~\cite{madaanrepresentational}. Similarly, MAS parameter regularization has been applied to MoCo-v2 \cite{chen2020mocov2} and BYOL \cite{grill2020bootstrap_latent} to close the performance gap between sequential and joint training on streaming data~\cite{huwell}. 
% The SSCFA framework~\cite{saratr} adapts MAS to regularize a Barlow-Twins loss for continual feature adaptation in Synthetic Aperture Radar (SAR) imagery. 

These methods have also been benchmarked in large-scale multimodal continual pretraining, where EWC and SI are frequently used as baselines~\cite{udandarao2024fomo_flux, zhu2023ctp, garg2024ticclip, yao2025retinal_vlp, ni2023modx, liu2025cclip}. However, empirical studies show that they underperform compared to alternative approaches. %, while incurring additional computational overhead.

\textbf{Takeaway:}
Although weight regularization may occasionally improve knowledge retention over other methods, these gains come at the cost of poor plasticity.
% However, these benchmarks reveal that while regularization aids knowledge retention, strictly penalizing weight updates can cripple model plasticity over long incremental trajectories. Therefore, while regularization effectively consolidates historical features, it must be carefully balanced to avoid restricting the network's capacity to absorb new distributions.

% In CBT \cite{marsocci2023cbt}, they continually train a Barlow Twins backbone on remote sensing data, and use EWC to mitigate forgetting. Other works also tried EWC as a baseline~\cite{gomez2022cl_pfr, fini2022cassle} and found that while it helps, it is typically outperformed by feature distillation approaches.
% In multimodal learning, standard weight regularization methods are also commonly used as baselines. However, across works, they also consistently underperform. 
% mod X used in vlp and synthetic

\subsection{Replay methods}
\label{sec:replay_methods}
Replay-based approaches maintain a memory buffer of past samples for reuse during the training of new tasks. The key differences between methods lie both in how these replay samples are used to mitigate forgetting and how they are selected and maintained over time.

A straightforward strategy is to directly concatenate some replay samples to each training batch. For instance, Singh et al. \cite{singh2025infcos} continually pre-trained an MAE in this manner using Experience Replay (ER) \cite{rolnick2019experience}, and observed low forgetting when combined with an infinite learning rate schedule. In multimodal models~\cite{liu2025vlm_cl_survey}, such as for example TiC-CLIP \cite{garg2024ticclip}, ER has also been shown to help significantly. In CTP \cite{zhu2023ctp} and RetCoP~\cite{yao2025retinal_vlp} they also observe that the performance of their method improves a lot when including replay. IncCLIP \cite{yan2022gntr_vlp} also uses ER, and additionally generates difficult pseudo-negative text samples for their contrastive loss through model inversion. Finally, in FoMo-in-Flux \cite{udandarao2024fomo_flux}, they experiment with an infinite buffer size in addition to access to the original pretraining pool, but they only allow finite compute. They show that in this setting, performance depends strongly on the training mixture: replaying buffered data is critical, replaying the pretraining pool has a limited effect, and increasing the ratio of replay samples versus new task data to make training more IID improves the accumulation-retention trade-off.

Rather than naively adding replay samples to each training batch, many methods incorporate replay samples through additional objectives, most commonly via distillation or contrastive losses. In CCL \cite{lin2022continual}, for example, replay samples are used both for distillation and for an auxiliary contrastive loss that separates representations of old and new tasks. Their buffer is constructed by selecting representative samples through K-Means clustering in embedding space. Similarly, MedCoSS \cite{ye2024medcoss} also adopts a K-Means-based buffer for feature distillation, and further applies MixUp \cite{zhang2018mixup} augmentation between samples of the same modality within the buffer.

A different line of work focuses on mixing replay and current samples via interpolation. A first work, LUMP \cite{madaanrepresentational}, uses MixUp to interpolate between samples from new tasks and the replay buffer. Similarly, Cro-Mo Mixup \cite{mushtaq2024cromo} adopts this same MixUp scheme and combines it with a Cross-Model Feature Mixup loss. CoSMAE \cite{mollenbrok2025cosmae} also incorporates this data mixup strategy alongside distillation. Finally, ASCL \cite{wu2023adaptive} also builds on LUMP, combining it with their adaptive sharpness-aware minimization (ASAM) to reduce overfitting, as well as their adaptive memory enhancement (AME), which applies augmentations of varying strength to increase buffer diversity.

Beyond how replay is used, several works investigate strategies for selecting informative samples. ASR \cite{cheng2023asr}, for example, selects a mixture of representative samples, whose representations remain stable under augmentation, and discriminative samples, whose representations vary the most. These are then used in their C\textsuperscript{2}ASR contrastive loss to maintain consistency with past tasks while separating them from new ones. CUCL \cite{cheng2023cucl} instead jointly trains their SSL objective along with additional codebooks for soft vector quantization of the representations. The samples whose representations are furthest away from their corresponding codewords are deemed the most difficult and informative, and are thus selected for replay. Finally, EDSR \cite{hanmo2024edsr} selects a small subset of replay samples per task by maximizing their Mutual Information with the full task dataset from an information-theoretic perspective. They also use these replay samples for distillation and mitigate potential overfitting on the replay buffer by making the replayed teacher outputs noisy.

In online CSSL settings, replay buffers serve multiple roles. First, they allow samples to be revisited multiple times instead of being observed only once. Second, they help shuffle and diversify the incoming data stream, which is particularly important when dealing with highly non-IID data such as video streams. Importantly, in this setting, sample selection must happen online, without access to the full task split and often without clear task boundaries, which requires strategies different from those used in offline CL.
In a work by Imai et al. \cite{imai2024faster}, for example, they keep a diverse buffer by retaining only samples with the lowest cosine similarity in feature space, reducing biased updates caused by non-IID data. In contrast, CLA \cite{cignoni2025clalatentalignmentonline} finds that a simple FIFO buffer performs better for their method. They argue that this makes the expected number of training iterations for each sample equal, which improves convergence and reduces overfitting on buffer samples, but this could come at the cost of a higher recency bias. Memory Storyboard \cite{yang2025memory} also uses a short-term FIFO buffer, but adds a larger long-term reservoir sampling memory on top of it to reduce such recency bias.

\textbf{Takeaway:}
Replay has often been shown to be an important component in CSSL pipelines. However, it should be noted that this comes at a cost. Old samples must be stored, which could cause problems in memory-constrained settings or when there are privacy concerns. Additionally, processing these extra replay samples can cost a significant amount of additional training compute.

\subsection{Architectural methods}

Architectural methods mitigate catastrophic forgetting at the parameter level by reducing interference between tasks. Rather than forcing a single shared set of weights to accommodate all tasks, these approaches freeze subsets of parameters or introduce task-specific components. This preserves previously learned representations while enabling controlled plasticity for new data.

In the context of CSSL, a representative example is Branch-Tuning \cite{Liu2025Branch}, which adds temporary, task-specific branches to existing network layers. During training on a new task, a parallel branch is added to the frozen backbone, and its outputs are combined. After training, these additional parameters can be incorporated into the original network, thus enabling knowledge integration without overwriting prior representations directly.

Some works adapt the model capacity online to match the evolving data distribution. For example, KIERA \cite{pratama2021kiera}  introduces a self-evolving deep clustering architecture, in which layers, nodes, and clusters are dynamically added during training. This enables the model to allocate new representational capacity to novel data while retaining previously learned structures.

Another approach exploits the hierarchical nature of deep networks by selectively freezing subsets of parameters. For example, PTLF \cite{yang2025PTLF} progressively freezes layers with high inter-task correlation, thereby reducing both forgetting and computational cost. This approach builds on the idea that initial representations in SSL remain consistent across tasks.

Finally, in vision-language continual learning, C-CLIP \cite{liu2025cclip} applies LoRA~\cite{hu2022lora} to both encoders of CLIP~\cite{CLIP2021Radford}, bounding weight deviation to limit forgetting but restricting plasticity in the process. FoMo-in-Flux \cite{udandarao2024fomo_flux} draws a similar conclusion: the number of trainable parameters strongly impacts the stability–plasticity trade-off, with parameter-selective methods showing strong retention but very limited accumulation, and parameter-efficient tuning approaches, such as LoRA, achieving a more favorable trade-off, but still exhibiting limited knowledge accumulation.

\textbf{Takeaway:}
Architectural methods reduce catastrophic forgetting by isolating parameters through freezing or improve plasticity by introducing task-specific modules. However, this often comes at the cost of limited scalability and reduced knowledge integration. Methods that freeze parts of the model prioritize stability over plasticity by restricting which parameters are updated, which makes adapting to new data more difficult. Furthermore, many methods depend on distinct task boundaries and may perform poorly in online settings. Overall, these methods demonstrate that freezing parameter updates is effective for retention and expanding the architecture improves plasticity, but both require careful design to balance flexibility and efficiency.

\subsection{Model merging}

Model merging has emerged as a computationally lightweight strategy to consolidate knowledge across fine-tuned models by directly aggregating their parameters. While most merging methods assume that all expert models are available simultaneously, continual learning introduces a sequential setting where models must be integrated over time.

In continual pretraining of CLIP models, TiC-CLIP~\cite{garg2024ticclip} evaluates a recursive merging strategy, where models are sequentially combined using interpolation coefficients tuned to preserve performance on previous tasks, but showed limited gains compared to replay-based approaches. FoMo-in-Flux~\cite{udandarao2024fomo_flux} shows that, together with replay, merging yields the strongest performance among evaluated methods. In particular, EMA merging and continual zero-shot merging achieve strong knowledge accumulation while maintaining neutral or improved zero-shot retention, especially over shorter training horizons. Building on these findings, TIME~\cite{dziadzio2025merge} provides a systematic analysis of temporal model merging, studying its use at both initialization and evaluation. They show that EMA-based strategies offer the best plasticity–stability trade-off, with limited sensitivity to the specific merging rule, and scale well across model sizes, compute budgets, and task sequence lengths.

\textbf{Takeaway:}
Model merging appears to be a promising and computationally efficient approach, including at scale, showing favorable behavior across both short and long update horizons, but has so far received limited attention in CSSL.

\subsection{Objective-level adaptation}
While most CSSL methods rely on mechanisms such as replay or distillation, a nascent line of work explores modifying the self-supervised objective itself to improve stability on non-stationary data streams, thereby reducing reliance on external components such as memory buffers or additional parameters in the form of teacher models.

A recent work, Continual Multi-Patch Learning (CMP)~\cite{cigoni2025ctp}, generates multiple patches from a single input and enforces consistency among their representations while preventing collapse through the Total Coding Rate (TCR) regularizer. Prior work by Imai et al. \cite{imai2024faster} used a similar setup in combination with replay, but CMP was shown to be effective even without requiring stored past samples, highlighting how objective design can substitute for explicit memory mechanisms.

In the spirit of tweaking the SSL objective, InfoUCL~\cite{zhang2024info_ucl} introduces an InfoDrop contrastive loss to mitigate the texture bias inherent in CSSL methods, improving generalization and reducing forgetting. Similarly, PNR~\cite{cha2024regularizing} mathematically redefines both contrastive and non-contrastive objectives to explicitly push new representations away from pseudo-negatives derived from the previous model. While both approaches embed continual learning constraints directly into the representation learning phase, they still fundamentally rely on external continual learning strategies.

\textbf{Takeaway:}
Objective-level adaptation represents a promising and underexplored direction for CSSL. By embedding CL constraints directly into the self-supervised objective, it offers an elegant pathway that bridges the two fields more tightly than approaches that address forgetting through external mechanisms.

% \begin{figure*}[t]
%     \centering
%     \includegraphics[width=\textwidth]{src/figures/fig18.png}
%     \caption{A simple image example}
%     \label{fig:cssl_example}
% \end{figure*}

\section{SSL as a component in hybrid pipelines}\label{sec:ssl_component}
% Kaizen, MAE-CIL, pMAE, DualNet
The robustness properties of SSL described in \Cref{sec:properties} suggest a natural question: can self-supervised objectives improve supervised continual learning, even when labels are available? Empirical evidence consistently answers yes. 

An early work by Zhang et al.~\cite{zhang2020ssl_aided_cil} showed that augmenting cross-entropy training with auxiliary self-supervised pretext tasks yields more generalizable, task-agnostic features that serve as a superior foundation for subsequent incremental classes. DualNet~\cite{pham2021dualnet} formalizes this intuition through a complementary learning system: a slow learner continuously optimizes a Barlow Twins objective to build general representations, while a fast learner adapts these for specific supervised classification tasks. Similarly, SscFA~\cite{dang2025sscfa} combines a similar architecture with a MAS weight regularization loss. Kaizen~\cite{tang2024kaizen} uses a setup similar to the fully self-supervised CaSSLe, training a self-supervised representation with replay and knowledge distillation. In addition, it adds a classifier on top. This classifier is trained during the task sequence whenever a sample has a label and also uses knowledge distillation and replay to mitigate forgetting.

MAEs have proven particularly versatile in hybrid settings as well, as their decoder can serve double duty: MAE-CIL~\cite{zhai2023mae_cil} exploits the MAE reconstruction to compress and decompress replay buffers, effectively increasing their replay buffer size without additional storage cost. In pMAE~\cite{he2024pmae}, a similar idea is used in federated CL using prompt tuning. 

Collectively, these works suggest that the benefits of SSL are not exclusive to fully unsupervised pipelines; they transfer meaningfully to supervised settings as well.

\section{Open challenges and future directions}\label{sec:discussion}
Despite rapid progress, CSSL is still in an early stage of development, with many core challenges remaining unresolved. While existing methods often demonstrate promising results on small-scale benchmarks, it is unclear whether these approaches can be applied to long-horizon or foundation model settings. A key issue is that much of the current literature adapts techniques from supervised CL without fully accounting for the unique properties of self-supervised objectives and representation learning. Consequently, fundamental questions remain regarding how to evaluate, scale, and control representation dynamics in non-stationary environments. We identify several critical challenges that define the current limitations of CSSL and outline directions for future research.

\subsection{Rethinking evaluation protocols and metrics}
A key limitation of current CSSL research is the lack of standardised evaluation protocols. As discussed in Section II, existing works differ substantially in terms of task construction (class-incremental versus stream-based), evaluation settings (offline versus online), and probe strategies (e.g., linear versus $k$-NN versus fine-tuning). This leads to inconsistent and often incomparable results. This variation makes it challenging to determine whether the reported improvements are due to methodological enhancements or variations in evaluation frameworks.

Furthermore, commonly used metrics such as linear probing or $k$-NN accuracy only offer a partial view of representation quality (linear separability and clustering), particularly in continual settings where adaptability, robustness to distribution shifts, and long-term retention are essential. This limitation is particularly evident in methods such as masked image modelling, where fine-tuning is a more accurate reflection of performance than probing~\cite{he2022masked_autoencoders}.

Crucially, forgetting and plasticity should also be analyzed from a purely representational perspective so they can be evaluated even under label scarcity. For example, one could measure uniformity and alignment~\cite{wang2020understanding} to track the structural health of representations over time, and monitor feature rank~\cite{garrido2023rankme} to assess whether the model maintains a high plasticity. 

Finally, we highlight the need for the development of unified, multidimensional evaluation protocols that assess representation quality, transferability, robustness, and efficiency consistently. Such benchmarks will be essential for driving meaningful progress and enabling CSSL to transition from controlled experimental setups to real-world deployment.

\subsection{Rethinking the stability–plasticity trade-off}
Although SSL objectives are more robust to forgetting than those of supervised learning, forgetting still accumulates over long training periods, particularly when there are significant shifts in the data distribution (see \Cref{sec:properties}). Consequently, much of the CSSL literature prioritizes stability over plasticity. The empirical observations reviewed in \Cref{sec:cssl} suggest that stability-prioritizing methods, such as weight regularization or naive feature distillation, often reduce plasticity by overly constraining representation updates.

However, a growing body of work challenges this framing directly: studies on projected distillation~\cite{gomez2022cl_pfr} and complementary network architectures~\cite{gomez2024plasticity} %, and remote sensing continual pretraining~\cite{mollenbrok2025plasticity} 
each argues that overly rigid alignment with past representations is itself a failure mode, one that prevents the model from integrating genuinely novel information. %The plasticity-aware remote sensing method~\cite{mollenbrok2025plasticity} goes further, concluding that for CSSL methods, effectively incorporating new knowledge should be treated as the primary objective, with forgetting prevention as a secondary constraint.
Therefore, we argue that plasticity should be regarded as a problem of equal importance to forgetting in CSSL.
This also opens a more precise research question: rather than uniformly minimizing forgetting, how can models distinguish between representations worth preserving and those that should be allowed to evolve, especially in unsupervised settings? Current methods treat all past knowledge as equally important, lacking mechanisms to identify transferable structure versus task-specific or obsolete features. Promising directions include parameter importance estimation conditioned on distribution shift, modular or compositional representations, and data-driven criteria for selectively updating features over time.

\subsection{Generalization of findings across SSL objectives}
A largely overlooked question in CSSL is whether empirical findings generalize across SSL objective families. As discussed in \Cref{subsec:ssl_objectives}, contrastive, non-contrastive, and masked modeling approaches differ in their stability properties, yet many CSSL studies commit to a single family per contribution (e.g. ~\cite{lin2022continual, gomez2022cl_pfr, mollenbrok2025cosmae}). Works that test across multiple SSL objectives, such as CaSSLe~\cite{fini2022cassle} and Branch-Tuning~\cite{Liu2025Branch}, reveal family-specific behaviors rather than uniform responses to the same CL strategy. CaSSLe observes that BYOL is unstable under naive fine-tuning while other methods are not, and that SwAV's prototype-based structure causes its representations to behave differently under $k$-NN probing compared to linear evaluation. Furthermore, distillation losses using L2-normalization conflict with Barlow Twins, which relies on standardization, leading to measurable performance degradation. In online settings, CLA~\cite{cignoni2025clalatentalignmentonline} additionally highlights that contrastive objectives are substantially more sensitive to minibatch size than non-contrastive ones, an important constraint when replay buffers are limited. Yet despite observing these interactions, neither CaSSLe nor Branch-Tuning systematically analyze why the same CL strategy produces different outcomes across families. Establishing which CSSL findings are objective-agnostic and which are family-specific is a prerequisite for building general-purpose continual pretraining approaches.

%\subsection{Scaling CSSL to foundation models and long-horizon settings}
\subsection{Scaling to foundation models and long-horizon settings}
% the field should move towards newer architectures

% Discuss about ViTs, trad SSL methods do not work that well with ViTs, maybe look into Dino and JEPA methods.
% Scalability with compute is important

A significant shortcoming of the existing literature is the discrepancy between experimental settings and real-world deployment scenarios. Except in multimodal settings, most CSSL methods are evaluated on small-scale benchmarks with simplified task constructions, and often rely on relatively small architectures such as ResNet-18 and datasets like CIFAR or TinyImageNet. There has been limited investigation into large-scale or foundation model regimes in unimodal vision settings. % This raises concerns about the data and architectural scalability of existing approaches. From the analysis in \Cref{sec:properties} and \Cref{sec:cssl}, it is clear that many current methods rely on components such as replay buffers, distillation networks, or additional objectives, which introduce significant computational and memory overhead.

This raises concerns about the data and architecture scalability of existing approaches. As discussed in \Cref{subsec:challenges} and \Cref{sec:cssl}, memory overfitting is a problem in CSSL, and at larger scales, the computational costs of current approaches could be prohibitive within constrained budgets. Even if compute were not a constraint, it is uncertain whether current solutions would improve with it. CLA~\cite{cignoni2025clalatentalignmentonline} is a concrete case where their approach does not scale with more compute, unless more replay data is used. Beyond raw compute, traditional SSL methods do not always translate seamlessly to ViT architectures~\cite{khan2025surveyself}. Future research must investigate how modern ViT-native self-supervised paradigms beyond CLIP, such as DINO~\cite{caron2021dino} or JEPA~\cite{assran2023jepa}, interact with continual data streams. Progress in this area will require developing parameter-efficient update mechanisms and compute-constrained strategies aligned with the specific challenges of modern foundation models.

Nonetheless, scaling is not only a computational challenge: it also exposes a deeper underspecification in current work. Deployed systems like the delivery robot from the introduction (\Cref{sec:intro}) would need to continually update their representations and knowledge to adapt to new environments or task requirements. However, most methods optimize for short stream benchmarks (see \Cref{subsec:benchmarks}) where retaining recent task performance is sufficient, without addressing how representations should evolve over genuinely long horizons, selectively discarding outdated features while preserving transferable structure. Closing this gap requires not just larger experiments, but theoretical frameworks that characterize what reliable, reusable representations look like under persistent non-stationarity.

Finally, another promising direction would look beyond label scarcity to address missing modalities in multimodal scenarios, establishing frameworks that maintain robust representations when a modality (e.g., text captions) is intermittently or permanently unavailable in new incoming training data.

% \section{Limitations}\label{limitations}
% \input{src/8_limitations}

\section{Conclusions}\label{conclusions}
This survey has examined CSSL for learning representations from unlabeled data in dynamic environments. We have reviewed training protocols, evaluation strategies, and methods for mitigating forgetting in sequential data streams. Although self-supervised objectives produce more general and stable representations than supervised approaches---a robustness we traced to task-agnostic features and flatter loss landscapes---they do not fully solve continual adaptation. Mechanisms such as projected feature distillation, replay, and architectural methods are still necessary, but raise scalability concerns that become more acute as model and data scale increase. Looking forward, we argue that progress requires moving beyond treating CSSL as supervised continual learning without labels. The lack of unified evaluation protocols and open questions around cross-family generalization of findings or the behavior of ViT-native objectives under non-stationary streams each demand approaches tailored to the specific properties of SSL. Addressing these challenges is essential to unlock the potential of CSSL as a backbone for lifelong foundation models that can adapt continuously to an ever-changing world.

% Despite progress, current methods, which are largely validated on small-scale benchmarks, fall short of the continual pretraining requirements of real-world applications. Closing this gap is essential to enable adaptive, lifelong foundation models in real-world applications.

\section*{Acknowledgements}
We would like to thank Emanuele Sansone and Alvaro Budria for their insightful comments. This paper has received funding from the Flemish Government under the Methusalem Funding Scheme (grant agreement n° METH/24/009).

\bibliographystyle{IEEEtran}
%\bibliography{references}
\bibliography{references_optimized} % checking how much space we can save by removing some URLs and abbreviating long conference names

@misc{bell2025futurecontinuallearningera,
      title={The Future of Continual Learning in the Era of Foundation Models: Three Key Directions}, 
      author={Jack Bell and others},
      year={2025},
      eprint={2506.03320},
      archivePrefix={arXiv},
      primaryClass={cs.LG},
      url={https://arxiv.org/abs/2506.03320}, 
}

@ARTICLE{jing2021selfsupervisedlearning,
author={Jing, Longlong and Tian, Yingli},
journal={ IEEE TPAMI },
title={{ Self-Supervised Visual Feature Learning With Deep Neural Networks: A Survey }},
year={2021},
volume={43},
number={11},
ISSN={1939-3539},
pages={4037-4058},
doi={10.1109/TPAMI.2020.2992393},
publisher={IEEE Computer Society},
address={Los Alamitos, CA, USA},
month=nov}

@article{kirkpatrick2017overcoming_forgetting,
  title={Overcoming catastrophic forgetting in neural networks},
  author={Kirkpatrick, James and others},
  journal={Proc. of the national academy of sciences},
  volume={114},
  number={13},
  pages={3521--3526},
  year={2017},
  publisher={National Academy of Sciences}
}

@article{CLIP2021Radford,
  author = {Radford, Alec and others},
  bibsource = {dblp computer science bibliography, https://dblp.org},
  description = {[2103.00020] Learning Transferable Visual Models From Natural Language Supervision},
  editor = {Meila, Marina and Zhang, Tong},
  eprint = {2103.00020},
  eprinttype = {arXiv},
  journal = {CoRR},
  pages = {8748-8763},
  title = {Learning Transferable Visual Models From Natural Language Supervision},
  url = {https://arxiv.org/abs/2103.00020},
  volume = 139,
  year = 2021
}

@article{shi2025cl_and_llms,
author = {Shi, Haizhou and others},
title = {Continual Learning of Large Language Models: A Comprehensive Survey},
year = {2025},
issue_date = {April 2026},
publisher = {Association for Computing Machinery},
address = {New York, NY, USA},
volume = {58},
number = {5},
issn = {0360-0300},
doi = {10.1145/3735633},
journal = {ACM Comput. Surv.},
month = nov,
articleno = {120},
numpages = {42}
}

@misc{oord2018info_nce,
      title={Representation Learning with Contrastive Predictive Coding}, 
      author={Aaron van den Oord and Yazhe Li and Oriol Vinyals},
      year={2019},
      eprint={1807.03748},
      archivePrefix={arXiv},
      primaryClass={cs.LG},
      url={https://arxiv.org/abs/1807.03748}, 
}

@inproceedings{chen2020sim_clr,
  title={A simple framework for contrastive learning of visual representations},
  author={Chen, Ting and Kornblith, Simon and Norouzi, Mohammad and Hinton, Geoffrey},
  booktitle={Int. Conf. on Machine Learning},
  pages={1597--1607},
  year={2020},
}

@inproceedings{He2019MomentumCF,
  title={Momentum contrast for unsupervised visual representation learning},
  author={He, Kaiming and Fan, Haoqi and Wu, Yuxin and Xie, Saining and Girshick, Ross},
  booktitle={Proc. of the IEEE/CVF Conf. on Computer Vision and Pattern Recognition},
  pages={9729--9738},
  year={2020}
}

@inproceedings{grill2020bootstrap_latent,
  title={Bootstrap your own latent-a new approach to self-supervised learning},
  author={Grill, Jean-Bastien and others},
  booktitle = {Advances in Neural Information Processing Systems},
  year= {2020},
  volume={33},
  pages={21271--21284},
}

@inproceedings{chen2022siamese_repr,
  title={Exploring simple siamese representation learning},
  author={Chen, Xinlei and He, Kaiming},
  booktitle={Proc. of the IEEE/CVF Conf. on Computer Vision and Pattern Recognition},
  pages={15750--15758},
  year={2021}
}

@InProceedings{zbontar2021barlow_twins,
  title={Barlow twins: Self-supervised learning via redundancy reduction},
  author={Zbontar, Jure and Jing, Li and Misra, Ishan and LeCun, Yann and Deny, St{\'e}phane},
  booktitle={Int. Conf. on Machine Learning},
  pages={12310--12320},
  year={2021},
  organization={PMLR}
}

@inproceedings{bardes2022vic_reg,
  author  = {Adrien Bardes and Jean Ponce and Yann LeCun},
  title   = {VICReg: Variance-Invariance-Covariance Regularization For Self-Supervised Learning},
  booktitle = {Int. Conf. on Learning Representations},
  year    = {2022},
}

@inproceedings{caron2020cluster_learning,
  title={Unsupervised learning of visual features by contrasting cluster assignments},
  author={Caron, Mathilde and others},
  journal={Advances in Neural Information Processing Systems},
  volume={33},
  pages={9912--9924},
  year={2020}
}

@inproceedings{he2022masked_autoencoders,
  title={Masked autoencoders are scalable vision learners},
  author={He, Kaiming and others},
  booktitle={Proc. of the IEEE/CVF Conf. on Computer Vision and Pattern Recognition},
  pages={16000--16009},
  year={2022}
}

@article{papyan2020prevalence_collapse,
  title={Prevalence of neural collapse during the terminal phase of deep learning training},
  author={Papyan, Vardan and Han, XY and Donoho, David L},
  journal={Proc. of the National Academy of Sciences},
  volume={117},
  number={40},
  pages={24652--24663},
  year={2020},
  publisher={National Academy of Sciences}
}

@inproceedings{tirer2023perturbation_collapse,
  title={Perturbation analysis of neural collapse},
  author={Tirer, Tom and Huang, Haoxiang and Niles-Weed, Jonathan},
  booktitle={Int. Conf. on Machine Learning},
  pages={34301--34329},
  year={2023},
  organization={PMLR}
}

@article{kornblith2021transfer,
  title={Why do better loss functions lead to less transferable features?},
  author={Kornblith, Simon and Chen, Ting and Lee, Honglak and Norouzi, Mohammad},
  journal={Advances in Neural Information Processing Systems},
  volume={34},
  pages={28648--28662},
  year={2021}
}

@article{masarczyk2023tunnel_effect,
  title={The tunnel effect: Building data representations in deep neural networks},
  author={Masarczyk, Wojciech and others},
  journal={Advances in Neural Information Processing Systems},
  volume={36},
  pages={76772--76805},
  year={2023}
}

@article{harun2024variables_generalization,
  title={What variables affect out-of-distribution generalization in pretrained models?},
  author={Harun, Yousuf and Lee, Kyungbok and Gallardo, Jhair and Krishnan, Giri and Kanan, Christopher},
  journal={Advances in Neural Information Processing Systems},
  volume={37},
  pages={56479--56525},
  year={2024}
}

@inproceedings{garrido2023rankme,
  title={Rankme: Assessing the downstream performance of pretrained self-supervised representations by their rank},
  author={Garrido, Quentin and Balestriero, Randall and Najman, Laurent and Lecun, Yann},
  booktitle={Int. Conf. on Machine Learning},
  pages={10929--10974},
  year={2023},
  organization={PMLR}
}

@inproceedings{caron2021dino,
  title={Emerging properties in self-supervised vision transformers},
  author={Caron, Mathilde and others},
  booktitle={Proc. of the IEEE/CVF Int. Conf. on Computer Vision},
  pages={9650--9660},
  year={2021}
}

@InProceedings{assran2023jepa,
    author    = {Assran, Mahmoud and others},
    title     = {Self-Supervised Learning From Images With a Joint-Embedding Predictive Architecture},
    booktitle = {Proc. of the IEEE/CVF Conf. on Computer Vision and Pattern Recognition},
    year      = {2023},
    pages     = {15619-15629}
}

@article{ozbulakknow,
  title={Know Your Self-supervised Learning: A Survey on Image-based Generative and Discriminative Training},
  author={Ozbulak, Utku and and others},
  journal={Transactions on Machine Learning Research},
  year={2024}
}

@inproceedings{huwell,
  title={How Well Does Self-Supervised Pre-Training Perform with Streaming Data?},
  author={Hu, Dapeng and others},
  booktitle={Int. Conf. on Learning Representations},
  year={2022}
}

@inproceedings{madaanrepresentational,
  title={Representational Continuity for Unsupervised Continual Learning},
  author={Madaan, Divyam and Yoon, Jaehong and Li, Yuanchun and Liu, Yunxin and Hwang, Sung Ju},
  booktitle={Int. Conf. on Learning Representations},
  year=2022
}

@inproceedings{bhat2022task,
  title={Task agnostic representation consolidation: a self-supervised based continual learning approach},
  author={Bhat, Prashant Shivaram and Zonooz, Bahram and Arani, Elahe},
  booktitle={Conf. on Lifelong Learning Agents},
  pages={390--405},
  year={2022},
  organization={PMLR}
}

@article{song2025discriminability,
  title={On the discriminability of self-supervised representation learning},
  author={Song, Zeen and Qiang, Wenwen and Zheng, Changwen and Sun, Fuchun and Xiong, Hui},
  journal={Information Sciences},
  pages={122556},
  year={2025},
  publisher={Elsevier}
}

@article{ben2023reverse,
  title={Reverse engineering self-supervised learning},
  author={Ben-Shaul, Ido and Shwartz-Ziv, Ravid and Galanti, Tomer and Dekel, Shai and LeCun, Yann},
  journal={Advances in Neural Information Processing Systems},
  volume={36},
  pages={58324--58345},
  year={2023}
}

@article{mollenbrok2025cosmae,
  title={Continual Self-Supervised Learning with Masked Autoencoders in Remote Sensing},
  author={M{\"o}llenbrok, Lars and Rasti, Behnood and Demir, Beg{\"u}m},
  journal={IEEE Geoscience and Remote Sensing Letters},
  year={2025},
  publisher={IEEE}
}

@article{singh2025infcos,
  title={Beyond Cosine Decay: On the effectiveness of Infinite Learning Rate Schedule for Continual Pre-training},
  author={Singh, Vaibhav and others},
  journal={Fourth Conf. on Lifelong Learning Agents},
  year={2025}
}

@inproceedings{ye2024medcoss,
  title={Continual self-supervised learning: Towards universal multi-modal medical data representation learning},
  author={Ye, Yiwen and others},
  booktitle={Proc. of the IEEE/CVF Conf. on Computer Vision and Pattern Recognition},
  pages={11114--11124},
  year={2024}
}

@inproceedings{zhai2023mae_cil,
  title={Masked autoencoders are efficient class incremental learners},
  author={Zhai, Jiang-Tian and Liu, Xialei and Bagdanov, Andrew D and Li, Ke and Cheng, Ming-Ming},
  booktitle={Proc. of the IEEE/CVF Int. Conf. on Computer Vision},
  pages={19104--19113},
  year={2023}
}

@inproceedings{he2024pmae,
  title={Masked autoencoders are parameter-efficient federated continual learners},
  author={He, Yuchen and Wang, Xiangfeng},
  booktitle={2024 IEEE Int. Conf. on Big Data (BigData)},
  pages={3682--3691},
  year={2024},
  organization={IEEE}
}

@inproceedings{yan2022gntr_vlp,
  title={Generative negative text replay for continual vision-language pretraining},
  author={Yan, Shipeng and others},
  booktitle={European Conf. on Computer Vision},
  pages={22--38},
  year={2022},
  organization={Springer}
}

@inproceedings{zhu2023ctp,
  title={Ctp: Towards vision-language continual pretraining via compatible momentum contrast and topology preservation},
  author={Zhu, Hongguang and Wei, Yunchao and Liang, Xiaodan and Zhang, Chunjie and Zhao, Yao},
  booktitle={Proc. of the IEEE/CVF Int. Conf. on Computer Vision},
  pages={22257--22267},
  year={2023}
}

@inproceedings{ni2023modx,
  title = {Continual Vision-Language Representation Learning with Off-Diagonal Information},
  author = {Ni, Zixuan and Wei, Longhui and Tang, Siliang and Zhuang, Yueting and Tian, Qi},
  booktitle = {Proc. of the 40th Int. Conf. on Machine Learning},
  year = {2023},
  pages = {26129--26149},
  publisher = {PMLR},
}

@inproceedings{liu2025cclip,
  title={C-CLIP: Multimodal continual learning for vision-language model},
  author={Liu, Wenzhuo and Zhu, Fei and Wei, Longhui and Tian, Qi},
  booktitle={Int. Conf. on Learning Representations},
  year={2025}
}

@inproceedings{fini2022cassle,
  title={Self-supervised models are continual learners},
  author={Fini, Enrico and others},
  booktitle={Proc. of the IEEE/CVF Conf. on Computer Vision and Pattern Recognition},
  pages={9621--9630},
  year={2022}
}

@inproceedings{gomez2022cl_pfr,
  title={Continually learning self-supervised representations with projected functional regularization},
  author={Gomez-Villa, Alex and Twardowski, Bartlomiej and Yu, Lu and Bagdanov, Andrew D and Van de Weijer, Joost},
  booktitle={Proc. of the IEEE/CVF Conf. on Computer Vision and Pattern Recognition},
  pages={3867--3877},
  year={2022}
}

@inproceedings{garg2024ticclip,
    title={TiC-CLIP: Continual Training of CLIP Models},
    author={ Garg, Saurabh and others},
    booktitle={Int. Conf. on Learning Representations},
    year={2024}
}

@inproceedings{yao2025retinal_vlp,
  title = {Continual Retinal Vision-Language Pre-training upon Incremental Imaging Modalities},
  author = {Yao, Yuang and Wu, Ruiqi and Zhou, Yi and Zhou, Tao},
  booktitle = {Medical Image Computing and Computer Assisted Intervention},
  year = {2025},
  pages = {111--121},
  publisher = {Springer}
}

@article{Liu2025Branch,
  title={Branch-tuning: balancing stability and plasticity for continual self-supervised learning},
  author={Liu, Wenzhuo and Zhu, Fei and Liu, Cheng-Lin},
  journal={IEEE Transactions on Neural Networks and Learning Systems},
  year={2025},
  publisher={IEEE}
}

@inproceedings{udandarao2024fomo_flux,
  title = {A Practitioner’s Guide to Continual Multimodal Pretraining},
  author = {Udandarao, Vishaal and others},
  booktitle = {Advances in Neural Information Processing Systems},
  year = {2024},
}

@misc{liu2025vlm_cl_survey,
  title = {Continual Learning for VLMs: A Survey and Taxonomy Beyond Forgetting},
  author = {Liu, Yuyang and others},
  year = {2025},
  eprint = {2508.04227},
  archivePrefix = {arXiv},
  primaryClass = {cs.CV},
  url = {https://arxiv.org/abs/2508.04227}
}

@inproceedings{van2018three,
  title={Three continual learning scenarios},
  author={Van de Ven, Gido M and Tolias, Andreas S},
  booktitle={NeurIPS Continual Learning Workshop},
  volume={1},
  number={9},
  pages={4},
  year={2018}
}

@article{ramaswamy2023geode,
  title={Geode: a geographically diverse evaluation dataset for object recognition},
  author={Ramaswamy, Vikram V and others},
  journal={Advances in Neural Information Processing Systems},
  volume={36},
  pages={66127--66137},
  year={2023}
}

@misc{ni2021selfsupervisedclassincrementallearning,
      title={Self-Supervised Class Incremental Learning}, 
      author={Zixuan Ni and Siliang Tang and Yueting Zhuang},
      year={2021},
      eprint={2111.11208},
      archivePrefix={arXiv},
      primaryClass={cs.LG},
      url={https://arxiv.org/abs/2111.11208}, 
}

@inproceedings{gomez2024plasticity,
  title={Plasticity-optimized complementary networks for unsupervised continual learning},
  author={Gomez-Villa, Alex and Twardowski, Bartlomiej and Wang, Kai and Van de Weijer, Joost},
  booktitle={Proc. of the IEEE/CVF Winter Conf. on Applications of Computer Vision},
  pages={1690--1700},
  year={2024}
}

@misc{cignoni2025clalatentalignmentonline,
      title={CLA: Latent Alignment for Online Continual Self-Supervised Learning}, 
      author={Giacomo Cignoni and Andrea Cossu and Alexandra Gomez-Villa and Joost van de Weijer and Antonio Carta},
      year={2025},
      eprint={2507.10434},
      archivePrefix={arXiv},
      primaryClass={cs.LG},
      url={https://arxiv.org/abs/2507.10434}, 
}

@inproceedings{marczak2024revisiting,
  title={Revisiting supervision for continual representation learning},
  author={Marczak, Daniel and Cygert, Sebastian and Trzci{\'n}ski, Tomasz and Twardowski, Bart{\l}omiej},
  booktitle={European Conf. on Computer Vision},
  pages={181--197},
  year={2024},
  organization={Springer}
}

@incollection{wu2023adaptive,
  title={Adaptive Self-Supervised Continual Learning},
  author={Wu, Lilei and Wang, Zhen and Liu, Jie},
  booktitle={ECAI 2023},
  pages={2680--2687},
  year={2023},
  publisher={IOS Press}
}

@article{cossu2024continual,
  title={Continual pre-training mitigates forgetting in language and vision},
  author={Cossu, Andrea and others},
  journal={Neural Networks},
  volume={179},
  pages={106492},
  year={2024},
  publisher={Elsevier}
}

@inproceedings{addepalli2022towards,
  title={Towards efficient and effective self-supervised learning of visual representations},
  author={Addepalli, Sravanti and Bhogale, Kaushal and Dey, Priyam and Babu, R Venkatesh},
  booktitle={European Conf. on Computer Vision},
  pages={523--538},
  year={2022},
  organization={Springer}
}

@inproceedings{koccyiugit2023accelerating,
  title={Accelerating self-supervised learning via efficient training strategies},
  author={Ko{\c{c}}yi{\u{g}}it, Mustafa Taha and Hospedales, Timothy M and Bilen, Hakan},
  booktitle={Proc. of the IEEE/CVF Winter Conf. on Applications of Computer Vision},
  pages={5654--5664},
  year={2023}
}

@inproceedings{imai2024faster,
  title={Faster Convergence and Uncorrelated Gradients in Self-Supervised Online Continual Learning},
  author={Imai, Koyo and Hayashi, Naoto and Hirakawa, Tsubasa and Yamashita, Takayoshi and Fujiyoshi, Hironobu},
  booktitle={Proc. of the Asian Conf. on Computer Vision},
  pages={436--453},
  year={2024}
}

@article{marsocci2023cbt,
  title={Continual barlow twins: continual self-supervised learning for remote sensing semantic segmentation},
  author={Marsocci, Valerio and Scardapane, Simone},
  journal={IEEE Journal of Selected Topics in Applied Earth Observations and Remote Sensing},
  volume={16},
  pages={5049--5060},
  year={2023},
  publisher={IEEE}
}

@inproceedings{zenke2017synaptic,
  title={Continual learning through synaptic intelligence},
  author={Zenke, Friedemann and Poole, Ben and Ganguli, Surya},
  booktitle={Int. Conf. on Machine Learning},
  pages={3987--3995},
  year={2017},
  organization={Pmlr}
}

@inproceedings{kornblith2019similarity,
  title={Similarity of neural network representations revisited},
  author={Kornblith, Simon and Norouzi, Mohammad and Lee, Honglak and Hinton, Geoffrey},
  booktitle={Int. Conf. on Machine Learning},
  pages={3519--3529},
  year={2019},
  organization={PMlR}
}

@inproceedings{lin2022continual,
  title={Continual contrastive learning for image classification},
  author={Lin, Zhiwei and Wang, Yongtao and Lin, Hongxiang},
  booktitle={2022 IEEE International conference on multimedia and expo (ICME)},
  pages={1--6},
  year={2022},
  organization={IEEE}
}

@misc{bagus2022supervisedcontinuallearningreview,
      title={Beyond Supervised Continual Learning: a Review}, 
      author={Benedikt Bagus and Alexander Gepperth and Timothée Lesort},
      year={2022},
      eprint={2208.14307},
      archivePrefix={arXiv},
      primaryClass={cs.LG},
      url={https://arxiv.org/abs/2208.14307}, 
}

@misc{kilickaya2023labelefficientincrementallearningsurvey,
      title={Towards Label-Efficient Incremental Learning: A Survey}, 
      author={Mert Kilickaya and Joost van de Weijer and Yuki M. Asano},
      year={2023},
      eprint={2302.00353},
      archivePrefix={arXiv},
      primaryClass={cs.LG},
      url={https://arxiv.org/abs/2302.00353}, 
}

@inproceedings{tang2024kaizen,
  title={Kaizen: Practical self-supervised continual learning with continual fine-tuning},
  author={Tang, Chi Ian and others},
  booktitle={Proc. of the IEEE/CVF Winter Conf. on Applications of Computer Vision},
  pages={2841--2850},
  year={2024}
}

@article{pham2021dualnet,
  title={Dualnet: Continual learning, fast and slow},
  author={Pham, Quang and Liu, Chenghao and Hoi, Steven},
  journal={Advances in Neural Information Processing Systems},
  volume={34},
  pages={16131--16144},
  year={2021}
}

@inproceedings{dziadzio2025merge,
  title={How to Merge Your Multimodal Models Over Time?},
  author={Dziadzio, Sebastian and others},
  booktitle={Proc. of the IEEE/CVF Conf. on Computer Vision and Pattern Recognition},
  pages={20479--20491},
  year={2025}
}

@article{
zhang2018mixup,
title={mixup: Beyond Empirical Risk Minimization},
author={Zhang, Hongyi and Cisse, Moustapha and Dauphin, Yann N. and Lopez-Paz, David},
journal={Int. Conf. on Learning Representations},
year={2018},
}

@inproceedings{mushtaq2024cromo,
  title={Cromo-mixup: Augmenting cross-model representations for continual self-supervised learning},
  author={Mushtaq, Erum and others},
  booktitle={European Conf. on Computer Vision},
  pages={311--328},
  year={2024},
  organization={Springer}
}

@inproceedings{aljundi2018mas,
  title={Memory aware synapses: Learning what (not) to forget},
  author={Aljundi, Rahaf and Babiloni, Francesca and Elhoseiny, Mohamed and Rohrbach, Marcus and Tuytelaars, Tinne},
  booktitle={Proc. of the European Conf on Computer Vision},
  pages={139--154},
  year={2018}
}

@article{dang2025sscfa,
  title={Self-Supervised Continual Learning for SAR-ATR: A Local Feature Adaptation Framework},
  author={Dang, Sihang and others},
  journal={IEEE Transactions on Aerospace and Electronic Systems},
  volume={62},
  pages={1378--1393},
  year={2025},
  publisher={IEEE}
}

@misc{zhang2020ssl_aided_cil,
      title={Self-Supervised Learning Aided Class-Incremental Lifelong Learning}, 
      author={Song Zhang and Gehui Shen and Jinsong Huang and Zhi-Hong Deng},
      year={2020},
      eprint={2006.05882},
      archivePrefix={arXiv},
      primaryClass={cs.LG},
      url={https://arxiv.org/abs/2006.05882}, 
}

@inproceedings{kilickaya2023labels,
  title={Are labels needed for incremental instance learning?},
  author={Kilickaya, Mert and Vanschoren, Joaquin},
  booktitle={Proc. of the IEEE/CVF Conf. on Computer Vision and Pattern Recognition},
  pages={2401--2409},
  year={2023}
}

@article{krizhevsky2009learning,
  author = {Krizhevsky, Alex},
  pages = {32--33},
  title = {Learning Multiple Layers of Features from Tiny Images},
  url = {https://www.cs.toronto.edu/~kriz/learning-features-2009-TR.pdf},
  year = 2009
}

@inproceedings{deng2009imagenet,
  title={Imagenet: A large-scale hierarchical image database},
  author={Deng, Jia and others},
  booktitle={2009 IEEE Conf. on Computer Vision and Pattern Recognition},
  pages={248--255},
  year={2009},
  organization={Ieee}
}

@inproceedings{cha2024regularizing,
  title={Regularizing with Pseudo-Negatives for Continual Self-Supervised Learning},
  author={Cha, Sungmin and Cho, Kyunghyun and Moon, Taesup},
  booktitle={Int. Conf. on Machine Learning},
  pages={6048--6065},
  year={2024},
  organization={PMLR}
}

@article{zhang2024info_ucl,
  title={InfoUCL: Learning informative representations for unsupervised continual learning},
  author={Zhang, Liang and Zhao, Jiangwei and Wu, Qingbo and Pan, Lili and Li, Hongliang},
  journal={IEEE Transactions on Multimedia},
  volume={26},
  pages={10779--10791},
  year={2024},
  publisher={IEEE}
}

@unknown{cigoni2025ctp,
      title={Replay-free Online Continual Learning with Self-Supervised MultiPatches}, 
      author={Giacomo Cignoni and Andrea Cossu and Alex Gomez-Villa and Joost van de Weijer and Antonio Carta},
      year={2025},
      eprint={2502.09140},
      archivePrefix={arXiv},
      primaryClass={cs.LG},
      url={https://arxiv.org/abs/2502.09140},
}

@inproceedings{pratama2021kiera,
  title={Unsupervised continual learning via self-adaptive deep clustering approach},
  author={Pratama, Mahardhika and Ashfahani, Andri and Lughofer, Edwin},
  booktitle={Int. Workshop on Continual Semi-Supervised Learning},
  pages={48--61},
  year={2021},
  organization={Springer}
}

@inproceedings{yang2025PTLF,
  title={Efficient self-supervised continual learning with progressive task-correlated layer freezing},
  author={Yang, Li and Lin, Sen and Zhang, Fan and Zhang, Junshan and Fan, Deliang},
  booktitle={26th Int. Symposium on Quality Electronic Design},
  pages={1--8},
  year={2025},
  organization={IEEE}
}

@inproceedings{hanmo2024edsr,
  title={Effective data selection and replay for unsupervised continual learning},
  author={Hanmo, LIU and others},
  booktitle={IEEE 40th Int. Conf. on Data Engineering},
  pages={1449--1463},
  year={2024},
  organization={IEEE}
}

@inproceedings{wang2020understanding,
  title={Understanding contrastive representation learning through alignment and uniformity on the hypersphere},
  author={Wang, Tongzhou and Isola, Phillip},
  booktitle={Int. Conf. on Machine Learning},
  pages={9929--9939},
  year={2020},
  organization={PMLR}
}

@article{rolnick2019experience,
  title={Experience replay for continual learning},
  author={Rolnick, David and Ahuja, Arun and Schwarz, Jonathan and Lillicrap, Timothy and Wayne, Gregory},
  journal={Advances in Neural Information Processing Systems},
  volume={32},
  year={2019}
}

@inproceedings{cheng2023asr,
  title={Contrastive continuity on augmentation stability rehearsal for continual self-supervised learning},
  author={Cheng, Haoyang and others},
  booktitle={Proc. of the IEEE/CVF Int. Conf. on Computer Vision},
  pages={5707--5717},
  year={2023}
}

@inproceedings{cheng2023cucl,
  title={CUCL: Codebook for unsupervised continual learning},
  author={Cheng, Chen and others},
  booktitle={Proc. of the 31st ACM Int. Conf. on Multimedia},
  pages={1729--1737},
  year={2023}
}

@article{yang2025memory,
  title={Memory Storyboard: Leveraging Temporal Segmentation for Streaming Self-Supervised Learning from Egocentric Videos},
  author={Yang, Yanlai and Ren, Mengye},
  journal={Fourth Conference on Lifelong Learning Agents},
  year={2025}
}

@misc{khan2025surveyself,
      title={A Survey of the Self Supervised Learning Mechanisms for Vision Transformers}, 
      author={Asifullah Khan and others},
      year={2025},
      eprint={2408.17059},
      archivePrefix={arXiv},
      primaryClass={cs.CV}
}

@inproceedings{hu2022lora,
  author  = {Hu, Edward and others},
  title   = {Lora: Low-rank adaptation of large language models},
  booktitle = {Int. Conf. on Learning Representations},
  year    = {2022},
}

@incollection{MCCLOSKEY1989109,
title = {Catastrophic Interference in Connectionist Networks: The Sequential Learning Problem},
editor = {Gordon H. Bower},
series = {Psychology of Learning and Motivation},
publisher = {Academic Press},
volume = {24},
pages = {109-165},
year = {1989},
issn = {0079-7421},
doi = {https://doi.org/10.1016/S0079-7421(08)60536-8},
author = {Michael McCloskey and Neal J. Cohen},

}

@misc{chen2020mocov2,
      title={Improved Baselines with Momentum Contrastive Learning}, 
      author={Xinlei Chen and Haoqi Fan and Ross Girshick and Kaiming He},
      year={2020},
      eprint={2003.04297},
      archivePrefix={arXiv},
      primaryClass={cs.CV},
      url={https://arxiv.org/abs/2003.04297}, 
}

\end{document}